\pdfoutput=1

\documentclass[11pt]{article}

\usepackage{EMNLP2023}

\usepackage{graphicx}
\usepackage{comment}
\usepackage{caption}
\usepackage{subcaption}
\usepackage{booktabs}
\usepackage{enumitem}
\usepackage{amsmath}
\usepackage{multirow}
\usepackage{diagbox}

\usepackage{times}
\usepackage{latexsym}

\usepackage[T1]{fontenc}

\usepackage[utf8]{inputenc}

\usepackage{microtype}

\usepackage{inconsolata}

\title{Uncovering the Full Potential of Visual Grounding Methods in VQA}

\author{Daniel Reich,  Tanja Schultz \\
         Cognitive Systems Lab., University of Bremen, Germany \\
         \texttt{\{dreich,tanja.schultz\}@uni-bremen.de}\\}

\begin{document}
\maketitle
\begin{abstract}
Visual Grounding (VG) methods in Visual Question Answering (VQA) attempt to improve VQA performance by strengthening a model's reliance on question-relevant visual information.  
The presence of such relevant information in the visual input is typically assumed in training and testing.
This assumption, however, is inherently flawed when dealing with imperfect image representations common in large-scale VQA, where the information carried by visual features frequently deviates from expected ground-truth contents. As a result, training and testing of VG-methods is performed with largely inaccurate data, which obstructs proper assessment of their potential benefits.

In this study, we demonstrate that current evaluation schemes for VG-methods are problematic due to the flawed assumption of availability of relevant visual information. Our experiments show that these methods can be much more effective when evaluation conditions are corrected.
Code is provided on GitHub\footnote{https://github.com/dreichCSL/TrueVG}.
\end{abstract}

\section{Introduction}
\label{sec:intro}
Visual Grounding (VG) in VQA has garnered interest not only as a key aspect to furthering understanding and rationalization of a VQA model's inference procedure, but also as a way to improve Out-of-Distribution (OOD) performance by preventing certain dataset biases to form. 
Various works have reported evidence of problematic tendencies in VQA models that point to a disregard of relevant image regions during answer inference and the manifestation of Q/A distribution biases in the model \cite{vqacp, goyal2017makingv, agrawal-etal-2016-analyzing_behavior}. A lack of VG in VQA models has been shown to negatively impact OOD performance \cite{reich2023fpvg} and have been tied to a general unpredictability of answering behavior \cite{Gupta2022SwapMixDA}. To alleviate these issues, methods have been developed that seek to strengthen a model's reliance on question-relevant visual features. 
These \textit{VG-methods} either modify the training procedure of existing models (e.g., HINT \cite{hint}, SCR \cite{Wu2019SelfCriticalRF}, VisFIS \cite{ying2022visfis}), or are integrated directly into specialized model architectures such as MMN \cite{chen2021meta}, PVR \cite{pvr} and VLR \cite{reich2022vlr}.

\begin{figure}[t]
\centering
\includegraphics[width=\columnwidth]{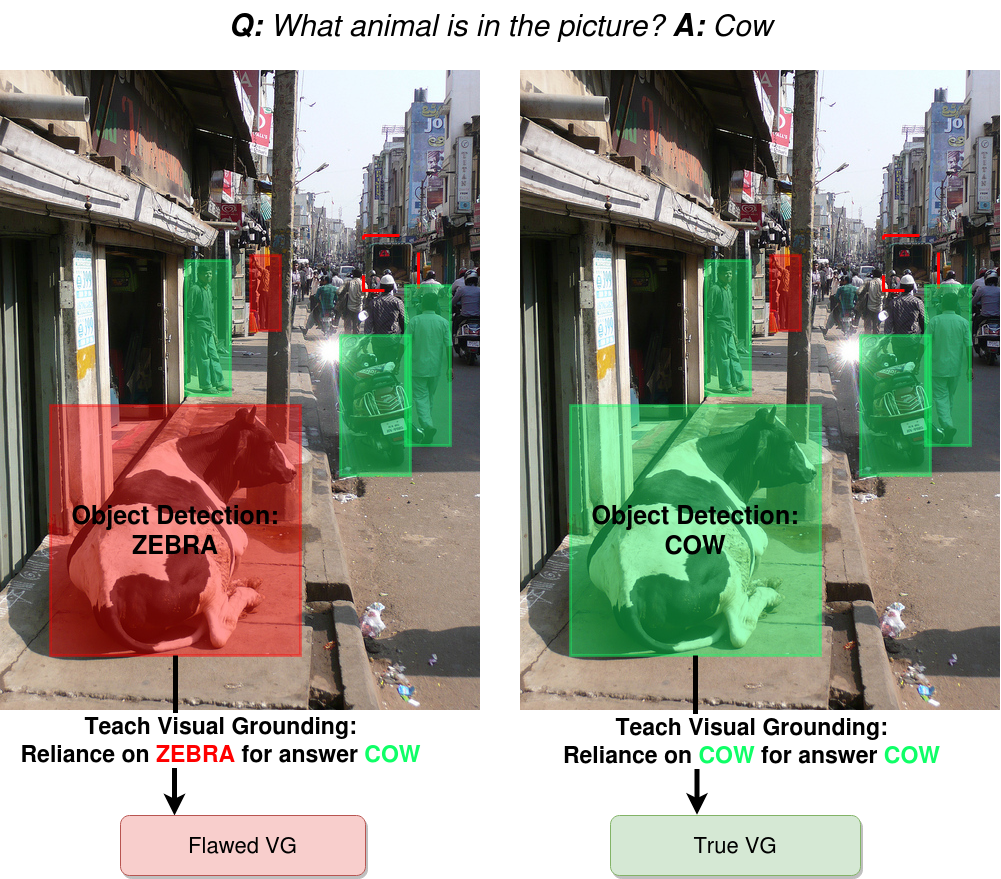}
\caption{Example of Flawed VG that VG-methods in VQA teach based on the unverified assumption of presence of relevant visual information (left). Correct content cues are a prerequisite for teaching True VG (right).}
\label{fig:cow_example}
\vspace{-4mm}

\end{figure}

On a technical level, the goal of VG-methods in VQA is to align a model's internal valuation of visual input feature importance (FI) with human-based FI, which is given as guidance in training. 
These Human-based FI scores can be inferred from a question's visual relevance annotations, which may be given as highlighted regions in the \textit{raw image} (e.g., spatial heat maps as in VQA-HAT \cite{Das2016HumanAI}), or explicit pointers to ground-truth objects (as in GQA \cite{gqa_dataset}). 
Notably, relevance annotations are not given in \textit{input feature space} directly, and therefore a mapping function is required to identify corresponding visual features and determine FI scores.
The predominant approaches for such a mapping between image and feature space rely exclusively on \textit{spatial matching}: 
Visual input features receive their FI scores depending on spatial overlap between the region they represent and annotated, question-relevant locations in the raw image (cf. \citet{hint, Wu2019SelfCriticalRF, shrestha-etal-2020-negative, ying2022visfis}). High-scoring features can then be identified as relevant cue objects\footnote{Visual feature vectors for VQA are commonly generated by object detectors such as Faster R-CNN \cite{fasterrcnn} and are therefore assumed to represent objects in an image.}.
In this approach, the actual \textit{visual content} carried by the cue objects is simply assumed to be appropriate without further \textit{semantic verification} and therefore does not influence their score. In this work, we report evidence that such incomplete verification can result in grossly mismatched cues, thereby leading to inadequate guidance in VG-method training, as illustrated in Fig. \ref{fig:cow_example}, left. Similarly, tests performed under such unchecked conditions fail to accurately evaluate the originally intended use case that VG-methods were designed for, as question-relevant content is often missing in the input and proper VG impossible. 
While work such as \citet{shrestha-etal-2020-negative} and \citet{ying2022visfis} investigate the underlying effects of VG-method application in detail, we are unaware of any study that also considers the impact of these problematic conditions in their analysis. 

In this study, we seek to develop a better understanding of the benefits of VG-methods in VQA when training and testing conditions properly support their intended use-case. We identify two flaws and their causes in current evaluation practices for VG-methods and outline an approach to fix them. Finally, we investigate their impact on VG-method training and testing in a series of experiments. The used methodology establishes a framework for evaluating VG-methods more thoroughly. 

\paragraph{Contributions.} Summarized as follows:
\begin{itemize}[noitemsep,nolistsep]
    \item An analysis of the flawed assumptions of cue object availability in current practices used for training and testing VG-methods in VQA.
    \item Comprehensive investigations of the impact of inadequate guidance for VG-method training.
    \item A methodology for training and testing VG-methods under corrected, proper conditions (code is provided).
\end{itemize}

\section{Related Work}
\label{sec:related_work}

\noindent\textbf{VG-Methods in VQA.} 
VG-boosting methods used for bias mitigation operate under the assumption that strengthening a model's reliance on relevant visual input will in turn weaken the influence of dataset-inherent biases towards Q/A priors and thereby improve OOD performance in VQA. Hence, evaluations on ID/OOD splits like \citet{vqacp} and \citet{ying2022visfis} are often used to evaluate their effectiveness for VQA. VG-boosting training-schemes may involve data augmentation with modulations of (relevant) visual input in image space \cite{mutant} or feature space \cite{Gupta2022SwapMixDA}, or training the model with objective functions that encourage an inference alignment with relevant image features for answer production, such as HINT \cite{hint} and SCR \cite{Wu2019SelfCriticalRF}. VisFIS \cite{ying2022visfis} combines both types of approaches in an ensemble of multiple objective functions.

\noindent\textbf{Relevance annotations and Feature Matching.}
VG-methods typically leverage annotations to point out question-relevant parts of the input image. VQA-HAT \cite{Das2016HumanAI} gathers such annotations in the form of spatial heat maps. These heat maps are recorded by tracking a user's computer mouse during de-blurring of image regions needed to answer crowd-sourced questions from the VQA dataset \cite{vqa}. \\
Template-based questions found in GQA \cite{gqa_dataset} and CLEVR-XAI \cite{clevr_xai} are generated in conjunction with an underlying visual scene graph and provide semantic relevance annotations of involved objects (or image regions) as a natural byproduct. \\
Computational approaches attempt to determine relevance annotations by employing a mapping between image annotations and question words \cite{mutant}, or by leveraging human-sourced textual explanations for the answer to a given question (VQA-X, \citet{Park2018MultimodalEJ_pointing_to_evidence}, used in \citet{Wu2019SelfCriticalRF}).

In all cases, the image-based relevance annotations are subsequently used to identify relevant cue objects in a model's visual input feature space. The mapping from annotations to input features has traditionally been based entirely on the features' receptive field, i.e., the image \textit{location} that the features represent \cite{hint, Wu2019SelfCriticalRF, shrestha-etal-2020-negative, ying2022visfis}. Our work goes one step further and examines VG-methods that consider cue objects that additionally match the \textit{content} of the relevant ground-truth object. To our knowledge, the effects of such \textit{semantic matching} on VG-method efficacy in VQA has not been explicitly investigated before.

\section{Flawed Visual Grounding}
\label{sec:flawed_visual_grounding}
We posit that evaluations of VG-boosting methods in VQA are following a flawed methodology, which consequently causes a flawed understanding of the benefits of VG and VG-boosting methods in VQA. \\
Specifically, we investigate the following issues:
\begin{enumerate}[noitemsep,nolistsep]
    \item[{(F1)}] \label{premise:testing} \textbf{Flawed testing:} Current evaluations of VG-methods hide their full potential by diluting tests with questions that are impossible to correctly ground due to missing relevant visual information.
    \item[{(F2)}] \label{premise:training} \textbf{Flawed training:} Impact of VG-boosting methods is muted due to training with a large percentage of unsuitable training samples that are missing relevant visual information necessary for teaching consistently correct inference alignments.
\end{enumerate} 

\noindent We identify two underlying \textit{causes} for these issues:
\begin{enumerate}[noitemsep,nolistsep]
    \item[{(C1)}] \textbf{Noisy features:} Impacts F1 and F2. Object misrecognitions and missing detections of relevant objects in the input image representation occur frequently in large-scale object detection tasks. We find that only 30\% of the used training questions (and 27\%/26\% of ID/OOD tests) in our GQA experiments contain all necessary question-relevant information. 
    \item[{(C2)}] \textbf{Fuzzy spatial matching of cue objects:} Impacts F2. Spatial identification of relevant cue objects may declare irrelevant objects as relevant on account of their close vicinity to the reference location, even if the represented visual content is inadequate and therefore irrelevant (as illustrated in Fig. \ref{fig:cow_example}). We identify a question-average of 2.6 cue objects in GQA training using semantic matching, which is inflated to 5.4 cue objects using spatial matching (counted based on a threshold of $IoU>0.5$; matching methods defined in Sec. \ref{sec:relevance_annotations}). This means that on average more than half of the objects that were declared question-relevant by spatial matching are in fact irrelevant. 
\end{enumerate}

\section{Experimental Setup}
We empirically show that addressing flaws F1 \& F2 outlined in Sec. \ref{sec:flawed_visual_grounding} provides new insights into the efficacy of VG-methods in VQA. In this section, we describe our methodology for fixing these flaws for analytical purposes.

\subsection{Approach}
\label{sec:approach}
\noindent\textbf{Enhancing the testing process (F1).} 
To better understand the potential of VG-methods we need to evaluate them on target cases where proper VG is at least feasible \textit{in principle}. A basic requirement for this is that question-relevant information needs to be fully represented in the visual input. 
Therefore, we determine ``True Visual Grounding'' (TVG) test subsets, which are verified to only contain questions that are accompanied by \textit{complete} relevant visual features (i.e., features that match all question-relevant reference annotations in both location \& content). 

\noindent\textbf{Enhancing the training process (F2).} 
VG-methods operate under the assumption that given training targets, i.e., visual features and their FI scores, are viable. Hence, training samples are expected to provide 1) relevant visual features carrying the content that is needed to answer the given question, and 2) FI scores that highlight them correctly in the set of all input features. Object detection-based visual features are noisy (see C1 in Sec. \ref{sec:flawed_visual_grounding}) and (parts of) the set of cue objects highlighted by spatial matching might be irrelevant and/or incomplete (see C2 in Sec. \ref{sec:flawed_visual_grounding}), thus resulting in failure to meet this requirement in many cases. 
We ensure availability of relevant visual content in the input by ``infusing'' missing information. These infused features can then be paired with perfect FI scores as guidance for VG-methods in training.

\subsection{From relevance annotations to cue objects}
\label{sec:relevance_annotations}

Relevance annotations point out the location (and in GQA also the identity) of ground-truth objects in the raw image that are relevant for answering a given question. 
In object-based VQA, the raw image and the actual image representation that is fed to the VQA model are not in the same space. Hence, a mapping is needed to identify cue objects in the input representation that match the relevance annotations. \\
We use two types of mapping methods:
\begin{enumerate}[noitemsep,nolistsep]
    \item[{(M1)}] \textbf{Spatial matching.} Cue objects are identified (scored) by measuring bounding box overlap (IoU) of (detected) visual input objects with relevant ground-truth objects. 
    \item[{(M2)}] \textbf{Semantic matching.} Scoring additionally involves verification that the \textit{content} of the spatially matched cue object matches the ground-truth object's identity (i.e., its name and attributes). 
\end{enumerate}

\noindent Typically, M1 is used in scoring relevant cue objects when relevance annotations point out relevant bounding boxes (such as in GQA). 
One of the reasons M2 has been neglected so far is the difficulty of interpreting sub-symbolic features w.r.t. their visual content. To circumvent this roadblock in our work, we engineer symbolic features that enable controlled and interpretable encoding of feature content (see Sec. \ref{sec:symbolic_features}). \\

\paragraph{Determining FI scores.}
Spatial and semantic matching are used to determine FI scores for each visual input object. These FI scores are subsequently used as guidance for VG-methods. \\
In \textit{spatial matching}, the FI score for each visual input object $o_{d}$ is set to the highest IoU match with any ground-truth question-relevant object $o_{gt}$, i.e., $s_{o_d} = max_{r\in GT} IoU(o_{d}, o_{gt}^r)$, where $GT$ is the set of question-relevant ground-truth objects. In this context, the calculated FI score $s_{o_d}$ can be interpreted as a measure of confidence that the object is relevant to the question, which is a reasonable way of compensating for the lack of insight into the object's actual informational content. \\
In \textit{semantic matching}, any detected object that meaningfully matches the location ($IoU>0.5$) \textit{and} fully matches the content of any ground-truth question-relevant object, is identified as question-relevant with full confidence. We therefore assign a maximum FI score to such input objects. Similarly, input objects that do not meet these requirements all receive a minimum FI score.

\subsection{Symbolic features}
\label{sec:symbolic_features}

\begin{figure}[h]
\centering
\includegraphics[width=1\columnwidth]{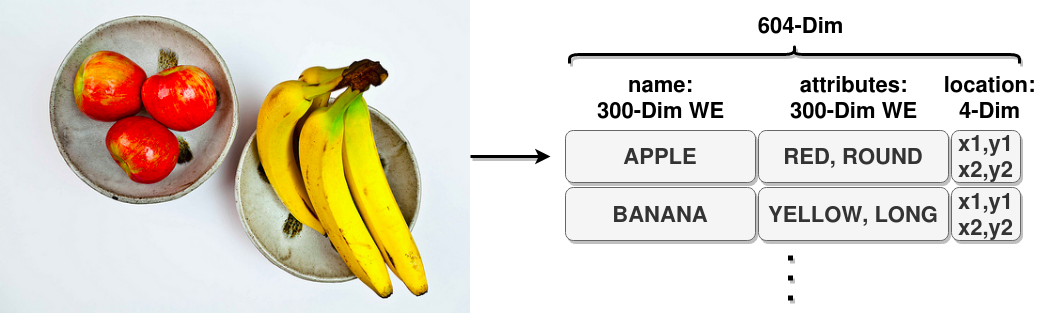}

\caption{Symbolic features.}
\label{fig:symbolic_features}
\vspace{-4mm}
\end{figure}

To gain a firm grasp on the informational content of the visual input, we engineer object-based \textit{symbolic} visual features instead of using standard \textit{sub-symbolic} features, which are commonly extracted from a late layer of an object detector like Faster R-CNN \cite{fasterrcnn}. 
Each symbolic feature vector represents an object in the image and carries information about its name, attributes and location. We encode each object with two stacked 300-dim GloVe word embeddings \cite{glove} and 4-dim for coordinates.
We construct two symbolic visual representations.  \\
\textbf{Detection (DET) features} are assembled from outputs of a scene graph generator. They represent the standard use case. \\
\textbf{Infusion (INF) features} are based on DET features, but are minimally ``infused'' with relevant, question-dependent information to enable what we call ``True VG'', i.e., training where the required feature content is present in the input and thus intended conditions for VG-methods are met. Note, that Infusion is applied in training only. \\
For each training question, we identify which relevant ground-truth objects are a) misrecognized (i.e., a meaningful spatial match ($IoU>0.5$) exists in DET but a semantic match does not), or b) missing entirely (no meaningful spatial match exists). Missing objects are introduced as new objects, assembled from image annotations. Misrecognized objects are adjusted to match the content of the corresponding ground-truth object (e.g., by replacing wrong object names and/or attributes with those given in annotations).
Additional details on these features and the used scene graph generator can be found in App. \ref{appsec:scene_graph_detection_and_feature_creation}.

\subsection{Used Datasets}
\label{sec:dataset}

\begin{table}[h]
\centering
\resizebox{\columnwidth}{!}{%
\begin{tabular}{lccccccc}
\toprule
 & & & \multicolumn{2}{c}{Test} && \multicolumn{2}{c}{True VG Subsets} \\
 \cmidrule{4-5} 
 \cmidrule{7-8}
Dataset & Train & Dev & ID & OOD && TVG-ID & TVG-OOD \\
\midrule
GQA-CP-large & 580k & 107k & 161k & 161k && 43k & 42k \\
VQA-HAT-CP & 32k & 6k & 4.1k & 5.9k && 1.1k & 1.6k \\
\bottomrule
\end{tabular}
}  
\caption{Sample counts for the used ID/OOD splits.}
\label{table:gqa_dataset}
\end{table}

Primary experiments are performed with the GQA dataset \cite{gqa_dataset}, which provides detailed scene graphs and semantic relevance annotations for most of its questions. Moreover, the GQA dataset uses template-generated questions that explicitly refer to information given in its scene graph annotations, thereby creating ideal conditions for our investigations. \\
In secondary experiments, we use the VQAv1-based \cite{vqa} VQA-HAT dataset \cite{Das2016HumanAI}, which provides relevance annotations as spatial heat maps over raw images without exact ties to specific ground-truth objects. The crowd-sourced commonsense-type questions in VQAv1 generally exhibit a much weaker connection to the image annotations than questions in GQA. 

Specifically, we use the two ID/OOD data splits GQA-CP-large and VQA-HAT-CP from \cite{ying2022visfis}, which were both created by a redistribution of questions following the ``changing priors'' approach used in the creation of the OOD split VQA-CP in \citet{vqacp}. 
We only train with questions that have meaningful relevance cue objects in all used visual feature types to achieve a fair comparison across model variants. Training set numbers listed in Table \ref{table:gqa_dataset} reflect this selection. On a similar note, all test sets for VQA-HAT-CP reflect the more challenging ``other''-type questions, as recommended in \citet{ood_testing_teney}.

\subsection{Used VQA Models}
\label{sec:vqa_models}

The classic attention-based, single-hop model UpDn \cite{bottomup_paper} traditionally takes center-stage in VQA's VG research. 
We additionally confirm GQA results with the more powerful transformer-based model LXMERT \cite{lxmert}. Results for LXMERT are reported in App. \ref{appsec:additional_evaluations_lxmert} and overall corroborate the insights gained for UpDn.
For training details see App. \ref{appsec:model_training_details}.

\subsection{Used VG-methods}
\label{sec:vg_methods}
We evaluate four VG-methods:
\begin{enumerate}[noitemsep,nolistsep]
    \item[{(1)}] AttAlign aligns a model's attention weights over visual input objects with the human-based FI scores. The model is trained by adding a cosine similarity-based loss to the standard VQA task loss.
    \item[{(2)}] HINT \cite{hint} aligns GradCAM \cite{gradcam} determined FI score rankings in the model with those given by human-based FI scores.
    \item[{(3)}] SCR \cite{Wu2019SelfCriticalRF} identifies a set of (ir)relevant objects by ranking human-based FI scores. The model is penalized if 1) an irrelevant object receives a higher GradCAM determined FI score (w.r.t. the ground-truth answer) than the top-scoring relevant object, and 2) if the top-scoring relevant object gets an even higher score for wrong answers.
    \item[{(4)}] VisFIS \cite{ying2022visfis} is a high-performing ensemble of ``Right for Right Reasons'' objectives, which includes cosine similarity-based FI alignment between human-based and model-based FI scores, as well as data augmentation based on (ir)relevant object sets determined by thresholding human-based FI scores. 
\end{enumerate}

\begin{table}[t]
\centering
\resizebox{0.85\columnwidth}{!}{%
\begin{tabular}{lccc}
\toprule
 \multicolumn{2}{c}{UpDn Training} & \multicolumn{2}{c}{Accuracy (All / TVG)} \\
 \cmidrule{3-4} 
 VG-method & Features & ID & OOD \\
\midrule
n/a & DET & 62.12 / 75.22 & 43.18 / 55.92 \\
 & INF & 61.16 / 78.78 & 45.40 / 62.54 \\
\midrule
VisFIS & DET  &  \textbf{62.51} / 76.83 & 44.06 / 57.60 \\
 & INF-spa  & 61.84 / 80.71 & 47.93 / 66.40 \\
 & INF-sem & 61.33 / \textbf{81.37} & \textbf{48.67} / \textbf{68.84} \\
\midrule
AttAlign & DET  & 61.18 / 74.63 & 42.30 / 55.25 \\
 & INF-spa  & 61.18 / 79.19 & 46.30 / 63.99 \\
 & INF-sem  & 60.74 / 80.12 & 46.84 / 66.17 \\
\midrule
HINT & DET  & 61.32 / 74.68 & 41.77 / 54.83 \\
 & INF-spa  & 61.37 / 79.40 & 46.41 / 63.43 \\
 & INF-sem  & 61.24 / 79.17 & 46.47 / 64.17 \\
\midrule
SCR & DET  & 61.84 / 75.16 & 42.89 / 56.13 \\
 & INF-spa  & 61.08 / 78.92 & 46.15 / 63.92 \\
 & INF-sem  &  61.28 / 79.18 & 46.63 / 64.34 \\
\bottomrule
\end{tabular}
}  
\caption{UpDn accuracies on GQA-CP-large.} 
\label{table:updn_results_small}
\vspace{-4mm}
\end{table}

\section{Impact on VQA Performance}
\label{sec:discussion}
Accuracies for UpDn models are listed in Table \ref{table:updn_results_small}. Note that all models are tested exclusively with DET features. We highlight certain results to illustrate the problem with Flawed VG.

\begin{figure}[t]
\large
\centering
\resizebox{\columnwidth}{!}{%
\includegraphics[width=\columnwidth]{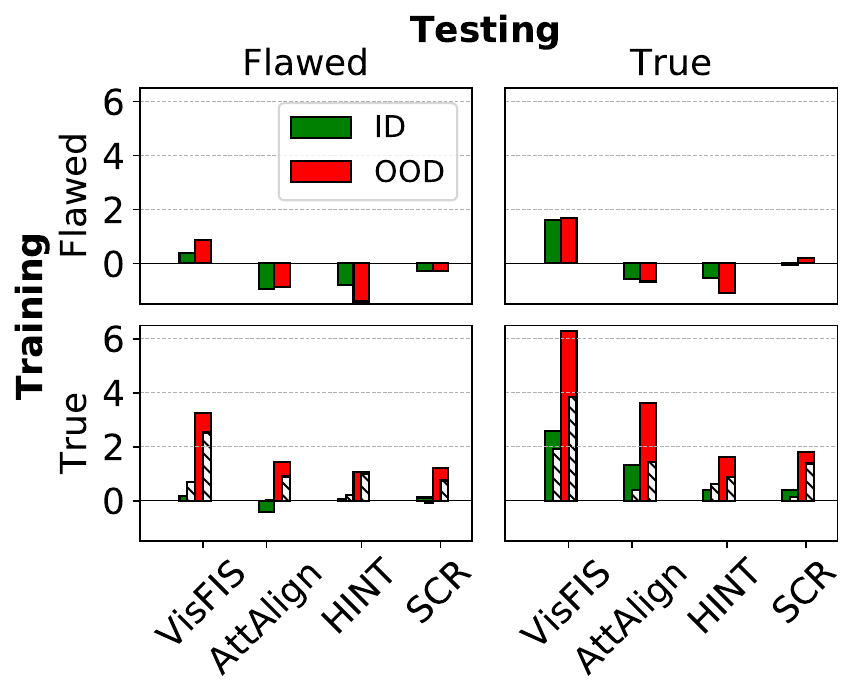}
}
\caption{Accuracy improvements from VG-methods compared to respective UpDn baselines. Training (y-axis): DET features with spatial matching (``Flawed'', top row), and INF features with semantic matching (``True'', bottom row). Striped bars: INF features with spatial matching. Testing (x-axis): Full test (``Flawed'') or TVG subset (``True'').}
\label{fig:updn_vgmethods}
\vspace{-4mm}
\end{figure}

\subsection{Flaw 1: Testing}
\label{subsec:flawed_testing}
We first evaluate impact of VG-methods on the True VG (TVG) tests, which only contain questions that are accompanied by complete relevant visual content.  
Fig. \ref{fig:updn_vgmethods} shows VQA performance impact on UpDn from VG-method training. The left column shows impact on full ID/OOD tests (labeled ``Flawed''), while the right column shows impact on TVG tests (``True'').
In comparison, Fig. \ref{fig:updn_vgmethods} shows that impact of VG-methods is considerably more pronounced in TVG testing (right column), confirming that Flaw 1 is indeed problematic for evaluating VG-methods, as their impact is significantly muted in testing. Moreover, these results suggest that VG-methods can be considerably more effective than previously shown.

\subsection{Flaw 2: Training}
\label{subsec:flawed_training}

\paragraph{Training with INF features and semantically-matched cues.}
Fig. \ref{fig:updn_vgmethods}, y-axis, shows accuracy improvements when applying VG-methods in training under ``Flawed VG'' and ``True VG'' settings. 
``Flawed'' training is performed with DET features and spatially matched cues (=standard case), while ``True'' training is performed with INF features and semantically matched cues (=intended case). 
Comparing top and bottom rows in Fig. \ref{fig:updn_vgmethods} reveals considerably greater impact of VG-methods in True VG training (bottom row), particularly for TVG tests in OOD. This shows that Flaw 2 is indeed a source of significant result distortion which may lead to misjudgement of a VG-method's efficacy.

\paragraph{Spatially vs. semantically matched cues.}
Striped bars in the bottom row in Fig. \ref{fig:updn_vgmethods} illustrate the results of ``True VG'' training using FI scores from spatial matching instead of semantic matching.
Differences between the two matching types are particularly noticeable in TVG testing (bottom right) and demonstrate that using semantically matched cues can amplify impact of VG-methods substantially.

\subsection{Flawed VG vs. True VG}
In summary, evaluations in ``Flawed VG'' settings (Fig. \ref{fig:updn_vgmethods}, top left) present rather weak evidence to suggest that VG-methods are particularly beneficial to VQA performance, while ``True VG'' (bottom right) reveals that they can in fact be \textit{very} effective when relevant visual information is present in the input. Particularly interesting here is the strong positive impact of the AttAlign method, which has been repeatedly declared ineffective in previous work, where conclusions were drawn based on the demonstrated flawed evaluations (cf. \citet{hint, ying2022visfis}).

\begin{figure*}
\begin{subfigure}[b]{0.46\textwidth}
    \includegraphics[width=\textwidth]{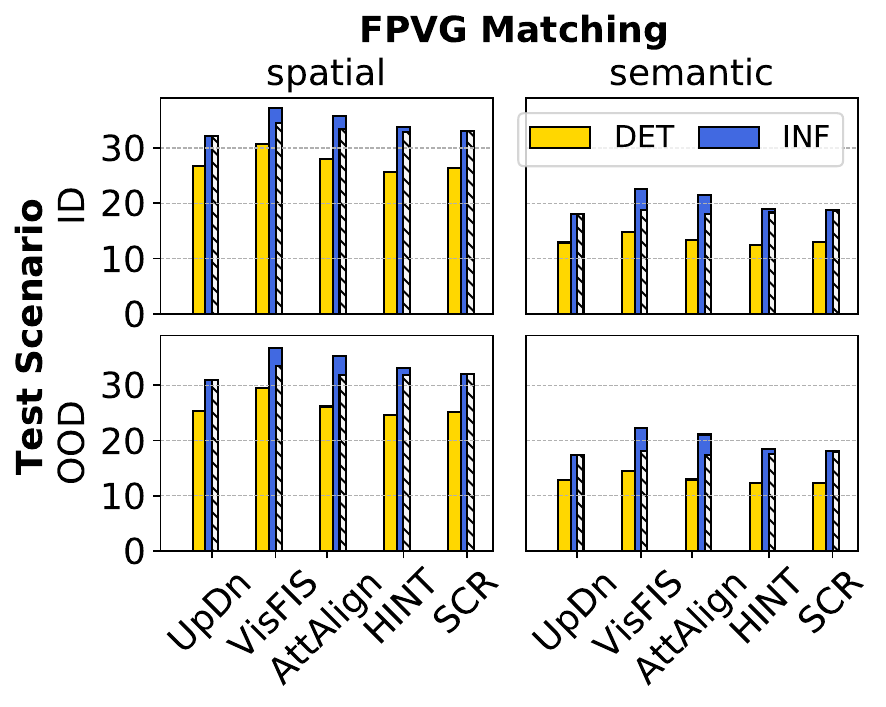}
\end{subfigure}
\begin{subfigure}[b]{0.46\textwidth}
    \includegraphics[width=\textwidth]{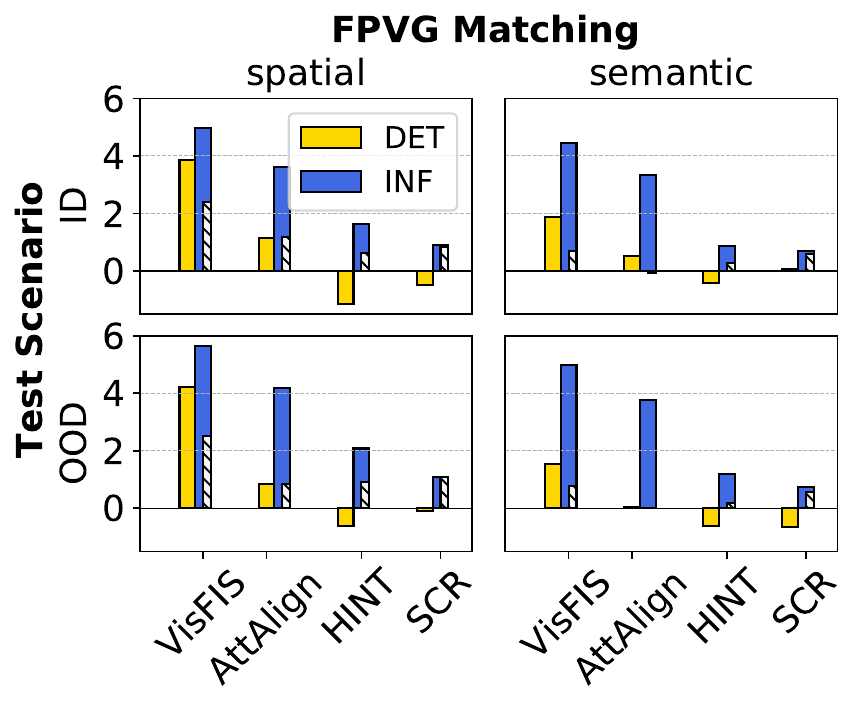}
\end{subfigure}
\hfill

 \vspace{-2mm}
      \caption{$FPVG_+$ measured on TVG subsets (ID/OOD) for UpDn. Left: Absolute $FPVG_+$ measurements. Right: $FPVG_+$ improvements compared to respective UpDn baselines. Columns categorize the matching method used for FPVG (see Sec. \ref{sec:vg_measurements}). Striped bars show results for INF-based models trained with spatial matching.}
     \label{fig:fpvg_updn_gqacp}
 \vspace{-4mm}

\end{figure*}

\section{Impact on Visual Grounding Quality}
\label{sec:vg_measurements}
Following recommendations in \cite{shrestha-etal-2020-negative}, we investigate impact of VG-methods on a model's Visual Grounding quality with the dedicated metric FPVG \cite{reich2023fpvg}.

\subsection{Relevance matching in FPVG}
FPVG measures a model's VG quality by confirming the model's reliance on question-relevant objects during answer inference. As with VG-method training, identifying question-relevant cue objects is a defining step in FPVG (which it is in any VG metric that aims to measure ``plausible'' VG, i.e., a model's VG w.r.t. visual objects deemed plausibly relevant to answer a given question).
The identification of relevant objects in FPVG follows the same procedure as the determination of FI scores for VG-method training (see Sec. \ref{sec:relevance_annotations}), i.e., it can be performed by spatial or semantic matching. FPVG was originally introduced with spatial matching on sub-symbolic visual features. Given that semantic matching provides the means of more accurately pinpointing the set of question-relevant (and irrelevant) objects in the visual input, we expect it to be able to support more precise VG measurements than location matching. We report FPVG results for both spatial and semantic matching, but surmise that semantic matching leads to a better-defined VG measurement in principle\footnote{We note that FPVG's goal is to quantify a certain desired manifestation of VG in VQA, which is described as a model's faithful reliance on plausibly question-relevant objects during answer inference (cf. \citet{reich2023fpvg}). It is vital to identify these objects correctly in order to produce a meaningful measurement in accordance with this goal.
Semantic matching leverages more available information from given relevance annotations to identify relevant input objects than spatial matching does. Therefore, we conclude that semantic matching provides more accurate object matches which enables more precise VG measurements with FPVG.}.

As only the TVG subsets provide full semantic matches, we discuss FPVG results on those subsets.

\subsection{Discussion}
Full numerical results for FPVG are listed in the Appendix in Table \ref{apptable:updn_results_all}. We highlight the most important results in Fig. \ref{fig:fpvg_updn_gqacp}.

\noindent\textbf{True VG training produces stronger VG than Flawed VG.}
Fig. \ref{fig:fpvg_updn_gqacp}, left, shows $FPVG_+$ measurements (i.e., the percentage of well-grounded questions in testing) in each evaluated UpDn model. 
True VG models (blue bars) achieve considerably higher levels of $FPVG_+$ than Flawed VG models (yellow bars) across all examined settings. This includes baseline models trained without VG-methods, which confirms that VG manifestation in models is generally held back when question-relevant information is not consistently provided.

\noindent\textbf{True VG training enables considerably higher $FPVG_+$ improvements than Flawed VG.}
Fig. \ref{fig:fpvg_updn_gqacp}, right, shows $FPVG_+$ improvements from training with VG-methods compared to respective UpDn baselines. Improvements are considerably stronger in True VG models (blue) than in Flawed VG models (yellow). This is especially true for FPVG with semantically matched targets (right column). \\
We interpret these results as additional evidence confirming the influence of Flaw 2 (muted impact of VG-methods due to Flawed VG training). In other words, VG-methods function more effectively in True VG training, where they can consistently align a model's inference with accurate and recurring (i.e., more stable) visual cue objects.

\noindent\textbf{Training with semantic matching has a considerably more positive effect on $FPVG_+$ than spatial matching.}
Striped bars in Fig. \ref{fig:fpvg_updn_gqacp} show $FPVG_+$ results for INF models trained with VG-methods and spatial matching. $FPVG_+$ is substantially more boosted with semantic matching (shown as blue bars). This illustrates that VG-methods can improve a model's VG quality more effectively when training with more accurate guidance.

\begin{figure*}
\begin{subfigure}[b]{0.33\textwidth}
    \includegraphics[width=\textwidth]{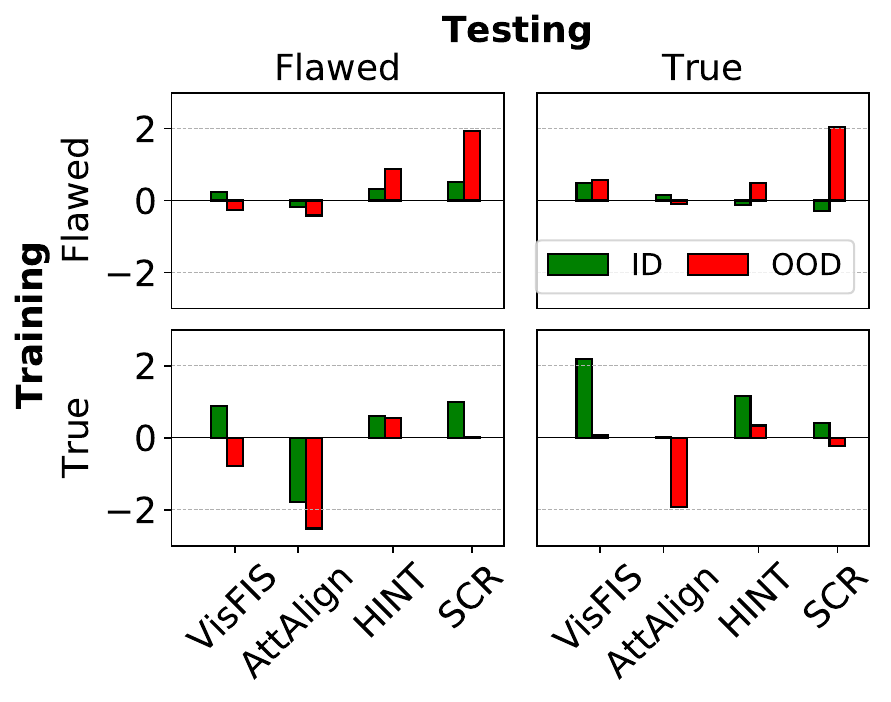}
\end{subfigure}
\begin{subfigure}[b]{0.33\textwidth}
    \includegraphics[width=\textwidth]{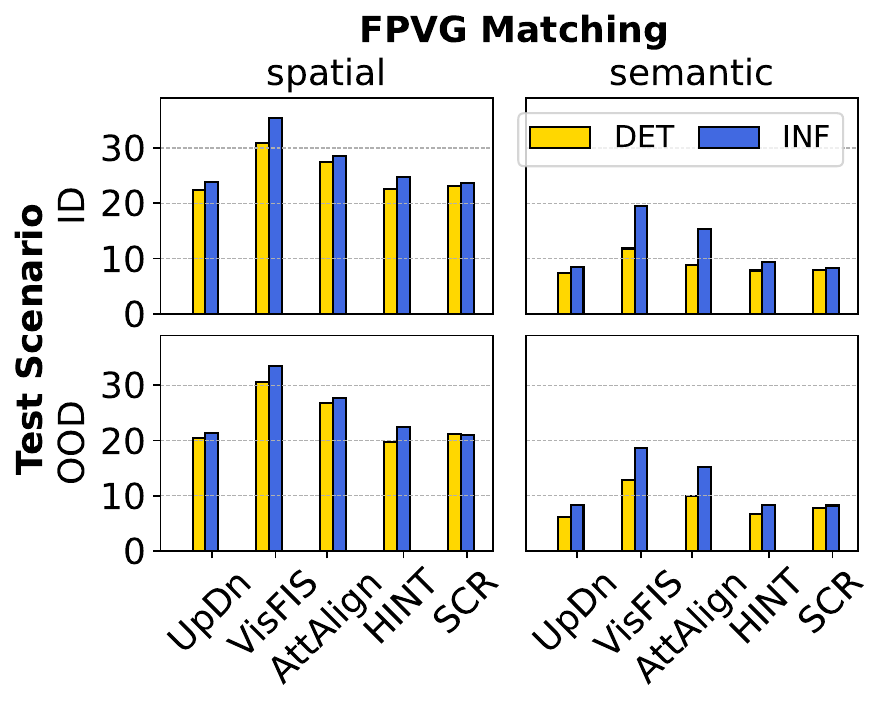}
\end{subfigure}
\hfill
\begin{subfigure}[b]{0.33\textwidth}
    \includegraphics[width=\textwidth]{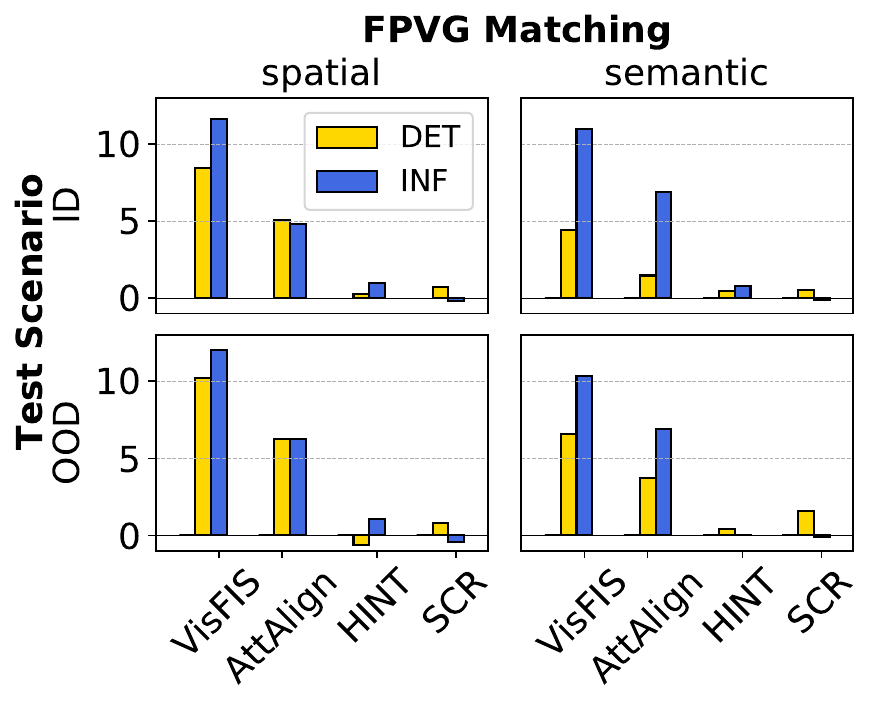}
\end{subfigure}
\hfill
 \vspace{-2mm}
      \caption{VQA-HAT-CP: Accuracy and $FPVG_+$ measurements (all values based on averages over five differently seeded UpDn models). See captions of Fig. \ref{fig:updn_vgmethods} and Fig. \ref{fig:fpvg_updn_gqacp} for a description of these histograms.}
     \label{fig:vqahat_threehistograms}
 \vspace{-2mm}

\end{figure*}

\section{True VG Analysis with VQA-HAT}
\label{sec:commonsense}

In this section, we adopt the True VG methodology for the VQAv1-based VQA-HAT-CP datset.

\subsection{Preliminaries}

\noindent\textbf{Model training.}
The UpDn training setup for VQA-HAT follows the settings used in \citet{ying2022visfis} (see also App. \ref{appsec:model_training_details}). 
We report averages and max deviation from the mean for five differently seeded training runs for each model variant.\\
\noindent\textbf{Visual features.}
We create symbolic features based on object detector outputs shared by \citet{bottomup_paper}, which provide object names and attributes for 36 objects per image. \\
\noindent\textbf{Evaluation.}
Accuracy calculations follow \citet{vqa}: A question is 100\% correct if the returned answer was given by at least 3 (of 10) annotaters and otherwise contributes a fractional score ($min(\frac{\#annotaters}{3}, 1)$) to overall accuracy.

\subsection{Dataset Challenges}
There are significant challenges to handle when evaluating VQA-HAT with True VG methodology.

\subsubsection{Sparse image annotations}
Infusion uses an image's object-level annotations to verify semantic matches between question-relevant objects and detected input objects. Furthermore, annotations act as source for infusing missing information. For VQA-HAT, we use the official MS COCO image annotations \cite{coco_dataset}, which is the underlying image database for the VQAv1 dataset. MS COCO annotates images with 80 different object names (GQA: 1702 classes). No attribute annotations are provided. 

\subsubsection{Relevance annotations}
\label{sec:relevance_annotations_vqahat}
VQA-HAT's relevance annotations are given as spatial heat maps, as opposed to GQA's unambiguous pointers to annotated objects in the image. 
As relevant ground-truth objects are not directly pointed out in VQA-HAT, but are required for Infusion, the first step here is to identify them in the image annotations corresponding to each heat map. 

A commonly used metric to determine object importance in VQA-HAT's heat maps was introduced in \citet{hint}. We adopt this metric to first calculate importance scores for all ground-truth objects in the image annotations and then apply a threshold to determine which objects are question-relevant. Concretely, the importance score for ground-truth object $o$ is calculated as $score_{o} = E_{in}(o) / (E_{in}(o)+E_{out}(o))$, where $E_{\{in,out\}}$ is the averaged pixel-level importance value inside and outside the ground-truth object's bounding box in the heat map. Following \citet{ying2022visfis}, we apply a threshold of 0.55 to these scores to establish the relevance status of each ground-truth object. 
Finally, FI scores for visual input features are determined with the same spatial or semantic matching procedure used for GQA (as described in Sec. \ref{sec:relevance_annotations}).

\subsection{Discussion}
\label{subsec:vqahat_discussion}
We highlight the most relevant results in Fig. \ref{fig:vqahat_threehistograms}. Full numerical results for VQA-HAT-CP are listed in the Appendix in Table \ref{apptable:vqahat_updn_results}. As with GQA, we investigate both accuracy and $FPVG_+$ changes for indicators that show how True VG evaluations improve our understanding of VG method impact.
Contrary to GQA, impact of True VG settings for VQA-HAT is considerably less pronounced and the observed trends are incongruous. For instance, strong improvements in $FPVG_+$ for VisFIS and AttAlign (Fig. \ref{fig:vqahat_threehistograms}, right) do not consistently translate to better accuracy (Fig. \ref{fig:vqahat_threehistograms}, left), as they did for GQA. Furthermore, differences in accuracy impact between full tests (=Flawed test case) and the TVG subsets (=True/intended test case) are  inconsistent across VG-methods, unlike for GQA, where they are strongly amplified for all VG-methods in True VG training (comparing bottom row in VQA-HAT's Fig. \ref{fig:vqahat_threehistograms}, left, with GQA's Fig. \ref{fig:updn_vgmethods}). \\
In summary, we observe no overarching, consistent benefits from adapting the True VG methodology to VQA-HAT, neither from a) introducing ground-truth object information in training, nor b) in testing cases where this annotated information is already present (TVG subsets). 
Definitive conclusions that explain these observations cannot be drawn due to the much less favorable dataset conditions compared to GQA, which hinder a proper True VG setup, combined with the different, more challenging nature of VQA-HAT's questions. 
VQA-HAT uses commonsense-type questions with less explicit references to relevant objects in the scene than GQA's retrieval-type questions. While the latter type can be expected to work well with symbolic features carrying object-centric information, the requirements for the former type's appropriate informational content is much less clear. This complicates the determination of what an optimal analytical setup for VG in VQA-HAT might entail.

In conclusion, we recommend performing True VG analysis with the GQA dataset, where the dataset conditions are more favorable and the VG task is better defined.

\section{Conclusion}
\label{sec:conclusion}
In this study, we have shown that current training and testing practices for VG-methods in VQA are flawed and therefore unable to reflect their full potential benefits for VQA models. We have proposed a methodology to optimize evaluation conditions to allow for a more thorough analysis of VG-method impact. Our investigations have shown that when conditions are optimized, VG-methods can elicit considerably stronger performance improvements in VQA models in both VG and accuracy, boosting OOD performance in particular. 

\section{Limitations}
\label{sec:limitations}
\paragraph{Transferability to other datasets.}
As demonstrated for VQA-HAT, transferability of the introduced methodology is conditioned on the availability of appropriate annotations and may be better suited to question types where the notion of what constitutes ``relevant content'' is well-defined and understood (such as in GQA). This limits the transferability of the methodology (and arguably poses a challenge for VG research in VQA as a whole). 

\paragraph{Usability of Infusion and semantic matching beyond analysis.}
Our experiments are based on non-standard symbolic features, which are not the first choice for image representations in current high-performing VQA models. While the evaluated fixes to Flawed VG in this work are leading to improvements in accuracy and VG and are useful for analysis of VG-methods, they cannot be applied to, e.g., higher performing symbolic features as is. We could imagine future work that investigates how feature content interpretation and Infusion can be applied to standard symbolic features to unshackle these ideas from its bond with symbolic features and make them useful beyond analysis.

\bibliography{TrueGrounding}
\bibliographystyle{acl_natbib}
\clearpage

\appendix

\section{Scene graph detection and symbolic feature creation}
\label{appsec:scene_graph_detection_and_feature_creation}

\subsection{Visual feature generation}
Symbolic features used in \textbf{GQA} evaluations were created based on classification outputs of the scene graph (SG) generator described in detail in \citet{reich2022vlr}. The used SG-detector uses a Faster R-CNN \cite{fasterrcnn} model for object detection for 1702 object classes. Attribute recognition for each detected object is done for each of the 39 attribute categories separately. Each category consists of two or more attribute classes (617 classes in total). Features for up to 100 objects are extracted per image.

For \textbf{VQA-HAT} evaluations and symbolic feature creation, we use the shared features generated with the object detector (OD) described in \citet{bottomup_paper}. The visual features from this OD provide the detected object class (one out of 1600 total) and up to one attribute class (out of 400 total) for 36 objects per image. 

\subsection{Symbolic feature creation}

The different types of symbolic features are created as follows:

\textbf{DET} features use the maximum class from object detection as object name (300D GloVe embedding) and the normalized 4D bounding box coordinates as location information. For determining attribute information (300D GloVe embedding), each category's maximum class is involved. To represent the attribute information of all categories in a single word embedding, we take the average of all recognized attribute names. \\
Note that in VQA-HAT evaluations we do not use bounding box information (following \citet{ying2022visfis}). Furthermore, whenever no attribute information was provided, a word embedding for the ``UNKNOWN'' token (the average over all word embeddings) was used. 

\textbf{ORA} (or Oracle) features for GQA are created accordingly, but without involvement of the SG-detector. These features are based on GQA's scene graph annotations which contain each object's name, attributes and location.

\textbf{INF} features use DET features as foundation. Modifications to object name information is realized by simple replacement. For attribute information modification in GQA evaluations, we first determine the attribute category of the annotated attribute that needs to be infused into an existing feature vector. Then, the embedding contributions of the (wrongly) recognized attribute of that category is replaced with an equivalently weighted embedding of the (correct) annotated attribute.

\section{Model Training Details}
\label{appsec:model_training_details}
For UpDn and LXMERT model training and evaluations, our experiments make use of implementations shared by \citet{ying2022visfis} (UpDn, LXMERT, VG-methods), \citet{lxmert} (LXMERT) and \citet{reich2023fpvg} (FPVG).

\subsection{UpDn}
UpDn models are trained for 50 epochs with 256 batch size for GQA-CP-large and 64 for VQA-HAT-CP. Model selection after 50 epochs is based on performance on the held-out dev set. We train either with or without one of the four examined VG-methods (VisFIS, AttAlign, HINT, SCR). Other hyperparameters, including training parameters specific to each VG-method were adopted from \citet{ying2022visfis}.

\subsection{LXMERT}

\textbf{Pre-training.} We pre-train LXMERT \cite{lxmert} from scratch for 30 epochs with 256 batch size, using all three types of symbolic features (DET, ORA, INF) to create three individual models. Model selection after 30 epochs is based on performance on the held-out dev set. We use the original implementation and training scheme described in the original paper, with a few changes: 
\begin{enumerate}[noitemsep,nolistsep]
    \item We exclude the attribute-related loss (unsuitable for more than a single attribute per object).
    \item We restrict pre-training to the GQA-CP-large train set (see Table \ref{table:gqa_dataset}) to retain ID/OOD distributional integrity in our experiments.
    \item Instead of the original setting of 36 objects, we use 100 visual objects (as provided by the used SG-detector).
    \item We use a smaller version of the model to speed up training and adapt to the reduced amount of training data: Hidden layer dimensions are reduced from 768 to 128. Intermediate layer size is reduced from 3072 to 512. Number of attention heads per self-attention layer is reduced from 12 to 4.
\end{enumerate}
Pre-training is done individually for each feature type (DET, ORA, INF). VG-methods are not applied in pre-training.

\textbf{Fine-tuning.} For fine-tuning, we train each model for 35 epochs with 64 batch size and use LXMERT's proposed two-layer classifier with softmax-based answer output. Model selection after 35 epochs is based on performance on the held-out dev set. 
Fine-tuning affects all LXMERT weights, not just the added VQA classifier.

\begin{table*}[t]
\centering
\resizebox{0.8\textwidth}{!}{%
\begin{tabular}{lcccccc}
\toprule
 \multicolumn{2}{c}{UpDn Training} & \multicolumn{2}{c}{Accuracy (All / TVG)} &&  \multicolumn{2}{c}{$FPVG_+$ (spatial / semantic)} \\
 \cmidrule{3-4} 
  \cmidrule{6-7} 
 VG-method & Features & ID & OOD && TVG-ID & TVG-OOD\\
 \midrule
 \midrule
n/a & DET  & 62.12 / 75.22 & 43.18 / 55.92 && 26.82 / 12.92 & 25.32 / 12.92 \\
 & ORA  & 55.46 / 71.84 & 46.88 / 63.78 && 32.85 / 18.59 & 32.68 / 19.01 \\
 & INF  & 61.16 / 78.78 & 45.40 / 62.54 && 32.19 / 18.07 & 31.03 / 17.35 \\
\midrule
 VisFIS & DET  & \textbf{62.51} / 76.83 & 44.06 / 57.60 && 30.68 / 14.81 & 29.52 / 14.47 \\
 & ORA  & 55.63 / 72.79 & 46.09 / 63.15 && 33.99 / 19.55 & 33.69 / 19.36 \\
 & INF-spa  & 61.84 / 80.71 & 47.93 / 66.40 && 34.59 / 18.77 & 33.55 / 18.14 \\
 & INF-sem  & 61.33 / \textbf{81.37} & \textbf{48.67} / \textbf{68.84} && \textbf{37.15} / \textbf{22.51} & \textbf{36.66} / \textbf{22.32} \\
\midrule
 AttAlign & DET  & 61.18 / 74.63 & 42.30 / 55.25 && 27.98 / 13.43 & 26.16 / 12.96 \\
 & ORA  & 54.55 / 71.40 & 45.49 / 62.52 && 32.59 / 18.64 & 31.29 / 18.69 \\
 & INF-spa  & 61.18 / 79.19 & 46.30 / 63.99 && 33.37 / 18.01 & 31.86 / 17.35 \\
 & INF-sem  & 60.74 / 80.12 & 46.84 / 66.17 && 35.79 / 21.42 & 35.21 / 21.10 \\
\midrule
 HINT & DET  & 61.32 / 74.68 & 41.77 / 54.83 && 25.67 / 12.51 & 24.69 / 12.31 \\
 & ORA  & 56.18 / 72.90 & 47.26 / 64.26 && 33.08 / 18.40 & 33.25 / 18.01 \\
 & INF-spa  & 61.37 / 79.40 & 46.41 / 63.43 && 32.83 / 18.34 & 31.94 / 17.53 \\
 & INF-sem  & 61.24 / 79.17 & 46.47 / 64.17 && 33.81 / 18.95 & 33.11 / 18.55 \\
\midrule
 SCR & DET  & 61.84 / 75.16 & 42.89 / 56.13 && 26.34 / 12.98 & 25.21 / 12.28 \\
 & ORA  & 56.00 / 72.67 & 47.26 / 64.52 && 32.82 / 18.57 & 32.26 / 18.41 \\
 & INF-spa  & 61.08 / 78.92 & 46.15 / 63.92 && 33.04 / 18.66 & 32.11 / 17.91 \\
 & INF-sem  & 61.28 / 79.18 & 46.63 / 64.34 && 33.10 / 18.76 & 32.11 / 18.09 \\
\bottomrule
\end{tabular}
}  
\caption{Complete UpDn results on GQA-CP-large.} 
\label{apptable:updn_results_all}
\vspace{-4mm}
\end{table*}


\section{Additional Evaluations}
\label{appsec:additional_evaluations}

\subsection{UpDn}
\label{appsubsec:additional_evaluations_updn}
All results for UpDn evaluated on GQA-CP-large are listed in Table \ref{apptable:updn_results_all}. Notably, these include additional evaluations with ORA features (creation thereof described in App. \ref{appsec:scene_graph_detection_and_feature_creation}).

\begin{figure}[h]
\large
\centering
\resizebox{\columnwidth}{!}{%
\includegraphics[width=1\columnwidth]{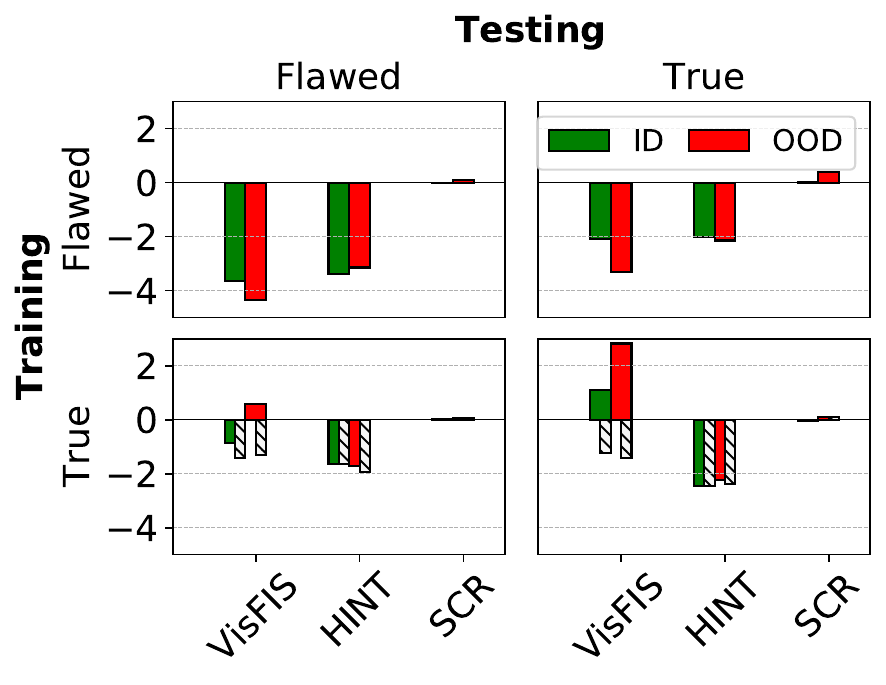}
}
\caption{Accuracy improvements for LXMERT+VG-methods compared to baseline LXMERT. Training (y-axis): DET features with location-matched cues (``Flawed'') or INF features with content-matched cues (``True''). Striped bars mark results when using location-matched cues instead. Testing (x-axis): Full test (``Flawed'') or True Grounding subset (``True'').}
\label{appfig:lxmert_vgmethods}
\end{figure}

\begin{figure*}[h]
\begin{subfigure}[b]{0.48\textwidth}
    \includegraphics[width=\textwidth]{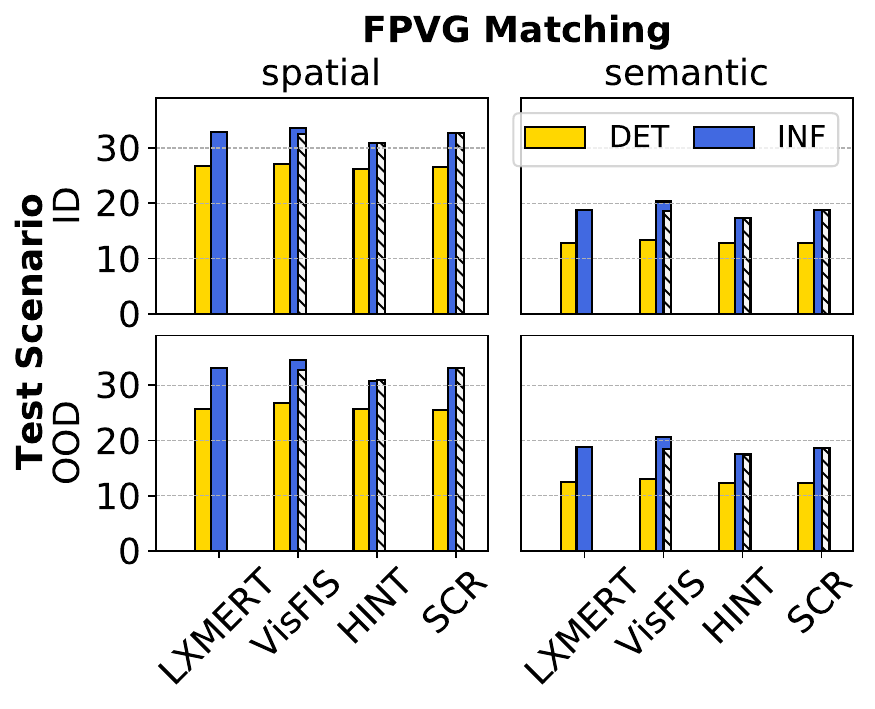}
\end{subfigure}
\begin{subfigure}[b]{0.48\textwidth}
    \includegraphics[width=\textwidth]{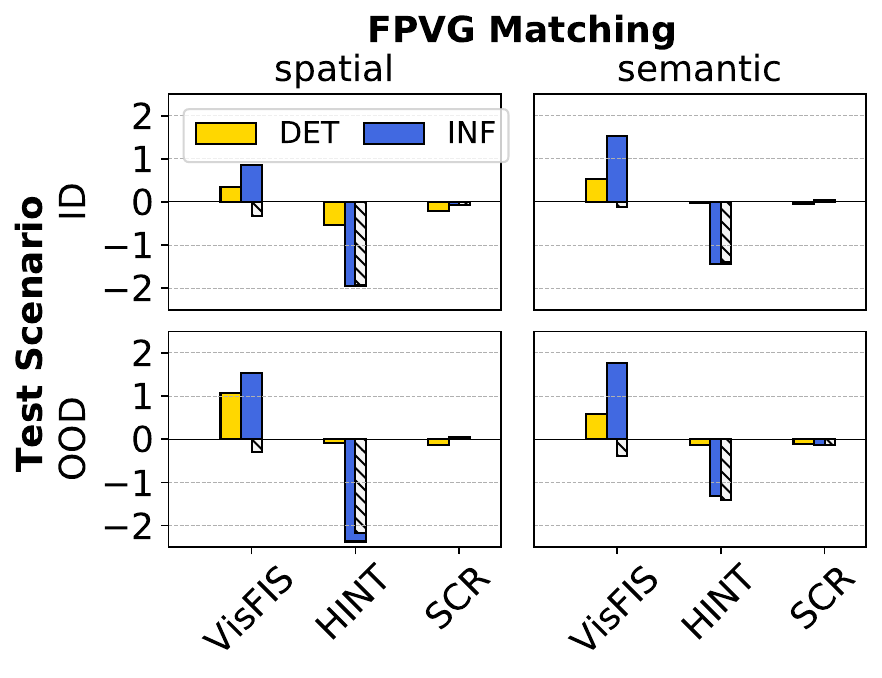}
\end{subfigure}
\hfill

 \vspace{-2mm}
      \caption{$FPVG_+$ measured on TVG subsets (ID/OOD) for LXMERT. Left: Absolute $FPVG_+$ measurements. Right: $FPVG_+$ improvements compared to respective UpDn baselines. Columns categorize the matching method used for FPVG (see Sec. \ref{sec:vg_measurements}). Striped bars show results for INF-based models trained with spatial matching instead of semantic matching.}
     \label{appfig:fpvg_lxmert_gqacp}
 \vspace{-2mm}

\end{figure*}

\begin{table*}[t]

\centering
\resizebox{0.8\textwidth}{!}{%

\begin{tabular}{lcccccc}
\toprule

 \multicolumn{2}{c}{LXMERT Training} & \multicolumn{2}{c}{Accuracy (All / TVG)} && \multicolumn{2}{c}{$FPVG_+$ (spatial / semantic)} \\
 \cmidrule{3-4} 
 \cmidrule{6-7}
 VG-method & Features &  ID & OOD &&  TVG-ID & TVG-OOD \\
 \midrule
 \midrule
n/a & DET  & \textbf{67.83} / 80.29 & 51.79 / 64.18 && 26.80 / 12.88 & 25.73 / 12.52 \\
 & ORA  & 58.22 / 75.45 & 51.42 / 69.36 && 33.25 / 19.59 & 34.53 / 20.25 \\
 & INF  & 64.95 / 82.75 & 52.69 / 69.90 && 32.83 / 18.80 & 33.10 / 18.86 \\
\midrule
 VisFIS & DET  & 64.19 / 78.21 & 47.43 / 60.87 && 27.14 / 13.41 & 26.80 / 13.10 \\
 & ORA  & 57.94 / 75.06 & 50.23 / 67.67 && 31.66 / 18.66 & 32.88 / 19.61 \\
 & INF-spa  & 63.55 / 81.52 & 51.37 / 68.47 && 32.50 / 18.69 & 32.81 / 18.48 \\
 & INF-sem  & 64.10 / \textbf{83.87} & \textbf{53.27} / \textbf{72.73} && \textbf{33.68} / \textbf{20.33} & \textbf{34.63} / \textbf{20.62} \\
\midrule
 HINT & DET  & 64.43 / 78.27 & 48.64 / 62.03 && 26.27 / 12.85 & 25.65 / 12.39 \\
 & ORA  & 58.01 / 74.95 & 51.07 / 68.63 && 32.15 / 19.22 & 33.81 /  20.10 \\
 & INF-spa  & 63.32 / 80.31 & 50.76 / 67.53 && 30.90 / 17.40  & 30.93 / 17.46 \\
 & INF-sem  & 63.32 / 80.30 & 50.97 / 67.66 &&  30.89 / 17.36 & 30.73 / 17.54 \\
\midrule
 SCR & DET  & 67.82 / 80.32 & 51.87 / 64.57 &&  26.60 / 12.84 &  25.59 / 12.41 \\
 & ORA  & 58.12 / 75.06 & 51.27 / 69.17 &&  32.95 / 19.47 & 34.11 / 20.29 \\
 & INF-spa  & 64.97 / 82.72 & 52.77 / 70.00 &&  32.76 / 18.85 &  33.14 / 18.73 \\
 & INF-sem  & 64.97 / 82.71 & 52.77 / 70.00 &&  32.76 / 18.85 & 33.14 / 18.73 \\
\bottomrule

\end{tabular}
}  
\caption{Complete LXMERT results on GQA-CP-large.} 
\label{apptable:lxmert_results_all}
\end{table*}

\subsection{LXMERT}
\label{appsec:additional_evaluations_lxmert}
Results for LXMERT on GQA-CP-large are listed in Table \ref{apptable:lxmert_results_all}. 
LXMERT evaluations that corroborate insights discussed for UpDn in Sec. \ref{sec:discussion} are illustrated in Fig. \ref{appfig:lxmert_vgmethods} (changes in accuracy) and Fig. \ref{appfig:fpvg_lxmert_gqacp} ($FPVG_+$ measurements), respectively. We summarize the following observations for LXMERT: 
\begin{enumerate}[noitemsep,nolistsep]
    \item Overall more favorable model behavior when training with VG methods in True VG settings (compared to Flawed VG) in both accuracy and $FPVG_+$.
    \item Strong negative impact to $FPVG_+$ when applying HINT in True VG settings, as well as a lack thereof in Flawed VG setting (see Fig. \ref{appfig:fpvg_lxmert_gqacp}, right, yellow vs. blue bars).
\end{enumerate}
W.r.t. (1): We observe no accuracy improvements for VisFIS under Flawed VG settings (see Fig. \ref{appfig:lxmert_vgmethods}, top row), while under True VG settings, improvements are made in particular in OOD settings (bottom row).
W.r.t. (2): While accuracy degradations for HINT in True VG and Flawed VG settings are comparable on the TVG subsets (see Fig. \ref{appfig:lxmert_vgmethods}, right column), we find that $FPVG_+$ measurements are expectedly negatively impacted (to a similar degree as accuracy) \textit{only} in True VG training (see Fig. \ref{appfig:fpvg_lxmert_gqacp}, right, blue vs. yellow bars). Similarly, observed VisFIS-driven accuracy improvements for True VG are accompanied by expected $FPVG_+$ improvements, while measurements for Flawed VG show contradicting readings. In summary, we observe a \textit{more congruent connection} between accuracy and VG under True VG settings.

\subsection{Oracle Features}
\label{appsubsec:oracle_features}
From a general performance standpoint, we investigate if ORA-features achieve similar results as INF-features without the added complexity involved in the latter's creation. 

In Table \ref{apptable:updn_results_all} and Table \ref{apptable:lxmert_results_all}, we find that both UpDn and LXMERT models trained with ORA-features generally exhibit weak ID accuracy compared to equivalently trained models with DET or INF features. In OOD testing, on the other hand, ORA-trained models perform surprisingly well, particularly when compared to DET-trained models. At the same time, we find considerable improvements in $FPVG_+$ for ORA-trained models, as well, albeit boosts gained from VG-methods are less pronounced than for DET and INF models. Apparently, ORA-trained models develop stronger grounding quality purely on account of complete and clear visual targets seen in training. The stronger grounding appears to positively affect their performance in OOD settings. These ORA results provide further evidence that VG is particularly important for OOD performance, as shown in \citet{reich2023fpvg}.


\begin{table*}
\centering
\resizebox{1\textwidth}{!}{%
\begin{tabular}{lcccccc}
\toprule
 \multicolumn{2}{c}{UpDn Training} & \multicolumn{2}{c}{Accuracy (All / TVG)} && \multicolumn{2}{c}{$FPVG_+$ (spatial / semantic)} \\
 \cmidrule{3-4}
  \cmidrule{6-7}
 VG-method & Features &  ID & OOD && TVG-ID & TVG-OOD \\
\midrule
\midrule
n/a & DET & 35.28$\pm$0.36 / 38.00$\pm$1.23 & 24.38$\pm$2.21 / 26.84$\pm$2.61 && 22.39$\pm$1.92 / 7.39$\pm$1.02 & 20.45$\pm$3.06 / 6.24$\pm$1.01 \\
 & INF & 35.24$\pm$0.56 / 37.80$\pm$1.07 & 25.65$\pm$2.06 / 28.56$\pm$2.27 && 23.80$\pm$2.40 / 8.52$\pm$1.54 & 21.44$\pm$4.02 / 8.37$\pm$1.50 \\
\midrule
VisFIS & DET & 35.52$\pm$2.89 / 38.49$\pm$3.71 & 24.13$\pm$2.94 / 27.40$\pm$3.00 && 30.82$\pm$11.99 / 11.83$\pm$3.25 & 30.66$\pm$9.68 / 12.79$\pm$2.33 \\
 & INF & 36.12$\pm$1.09 / \textbf{39.99}$\pm$2.12 & 24.87$\pm$1.49 / 28.62$\pm$1.68 && \textbf{35.46}$\pm$3.97 / \textbf{19.48}$\pm$1.74 & \textbf{33.47}$\pm$3.64 / \textbf{18.73}$\pm$2.82 \\
\midrule
AttAlign & DET & 35.10$\pm$1.25 / 38.14$\pm$0.77 & 23.97$\pm$2.14 / 26.74$\pm$1.90 && 27.46$\pm$1.45 / 8.86$\pm$1.43 & 26.72$\pm$1.96 / 9.94$\pm$1.97 \\
 & INF & 33.46$\pm$1.38 / 37.83$\pm$2.05 & 23.13$\pm$2.33 / 26.63$\pm$2.85 && 28.61$\pm$3.22 / 15.38$\pm$3.45 & 27.67$\pm$2.26 / 15.25$\pm$2.46 \\
\midrule
HINT & DET & 35.61$\pm$0.66 / 37.88$\pm$0.86 & 25.26$\pm$1.96 / 27.33$\pm$2.44 && 22.63$\pm$2.74 / 7.85$\pm$2.32 & 19.81$\pm$3.87 / 6.68$\pm$2.02 \\
 & INF & 35.85$\pm$0.34 / 38.95$\pm$0.58 & 26.21$\pm$0.98 / \textbf{28.90}$\pm$1.82 && 24.77$\pm$2.40 / 9.34$\pm$1.21 & 22.49$\pm$3.02 / 8.42$\pm$1.03 \\
\midrule
SCR & DET & 35.81$\pm$1.26 / 37.71$\pm$1.43 & \textbf{26.32}$\pm$1.87 / 28.88$\pm$1.37 && 23.09$\pm$2.22 / 7.94$\pm$1.04 & 21.23$\pm$2.70 / 7.84$\pm$2.11 \\
 & INF & \textbf{36.23}$\pm$0.51 / 38.22$\pm$0.94 & 25.66$\pm$1.80 / 28.34$\pm$1.96 && 23.59$\pm$1.66 / 8.38$\pm$0.99 & 20.98$\pm$2.96 / 8.27$\pm$1.46 \\
\bottomrule
\end{tabular}
}  
\caption{Accuracies and $FPVG_+$ measurements for UpDn evaluated on VQA-HAT-CP (only ``other''-type questions). We report average results and maximum deviation over five differently seeded training runs per model variant.}
\label{apptable:vqahat_updn_results}
\end{table*}

\end{document}